\documentclass[letterpaper, 10 pt, conference]{ieeeconf}  

\IEEEoverridecommandlockouts                              

\usepackage{graphics} 
\usepackage{epsfig} 
\usepackage{mathptmx} 
\usepackage{times} 
\usepackage{amsmath} 
\usepackage{amssymb}  
\usepackage{url}

\title{\LARGE \bf
TacHair: Tactile Contact-Distribution Guided Online Correction for Robotic Hair Stroking and Perception
}

\author{
Ruiyi Hu$^{1,*}$,
Yongqiang Zhao$^{1,*}$,
Daniel Bak$^{1}$,
Yupeng Wang$^{1}$,
Xuyang Zhang$^{1}$,
Shan Luo$^{1}$%
\thanks{This work was supported by the UKRI EPSRC project
``TacDiff: Designing Tactile-based Robots via Differentiable
Simulations'' (UKRI3438).}%
\thanks{$^{1}$Ruiyi Hu, Yongqiang Zhao, Daniel Bak, Yupeng Wang,
Xuyang Zhang, and Shan Luo are with King's College London,
London WC2R 2LS, United Kingdom.
Email: \texttt{yongqiang.zhao@kcl.ac.uk}.}%
\thanks{$^{*}$Contributed equally to this work.
Correspondence to Yongqiang Zhao.}%
}

\usepackage{graphicx}
\begin{document}

\maketitle
\thispagestyle{empty}
\pagestyle{empty}

%

\begin{abstract}
Hair stroking is common in daily grooming and personal care, and is also widely used in hair-product evaluation, motivating robots with similar physical interaction capabilities. Existing robotic hair-care and surface-following methods mainly rely on trajectory planning, compliance, force regulation, or tactile-conditioned policies, but deformable hair can remain in contact while gradually drifting across the end-effector, making local interaction difficult to regulate. We propose TacHair, a tactile contact-distribution guided online correction framework that represents high-resolution tactile observations as a spatial hair-contact distribution. A visuotactile imitation policy generates the nominal stroking motion, while a separately trained residual module corrects local contact deviations, separating task progression from contact recovery. We evaluate TacHair in 525 real-robot trials across five head geometries and three hair conditions. A successful stroke requires both sufficient task progression and contact maintenance; our method improves success from 42.9\% to 62.3\% and contact maintenance from 59.4\% to 88.0\% over the same visuotactile policy without correction. These results demonstrate spatial tactile contact distributions as an effective feedback representation for contact-preserving interaction with deformable and visually occluded surfaces. Demos, code, and datasets are available at \url{https://tachair.github.io}.
\end{abstract}


\section{Introduction}
\label{sec:introduction}

Hair stroking is a common physical interaction in daily grooming, personal care, and interpersonal assistance. Similar stroking and combing motions are also widely used in hair-product evaluation and hair-condition assessment, where the interaction between an applicator and hair provides information about properties such as smoothness, friction, and combability~\cite{suzuki1999combing,bhushan2005hairfriction}. As robots increasingly enter domestic, assistive, and product-development environments, it is therefore desirable for them to acquire comparable capabilities for sustained and adaptive physical interaction~\cite{billard2019trends,nanavati2024physically}. Unlike conventional pick-and-place manipulation, hair stroking requires the robot to continuously progress along a curved surface while maintaining an appropriate local interaction with highly deformable fibers. The difficulty lies not only in reaching the desired region, but also in regulating contact throughout the entire motion as the interaction interface evolves.

Existing robotic hair-care systems address related problems mainly through visual trajectory planning, compliant mechanisms, and force feedback. Prior work has demonstrated autonomous hair combing using hair segmentation and path planning~\cite{dennler2021haircombing}, adaptive brushing based on interaction forces~\cite{hughes2021detangling}, compliant head-care systems for washing and massage~\cite{hirose2013headcare}, and soft end-effectors for contact-rich tasks such as head patting and finger combing~\cite{yoo2025moehair}. In parallel, tactile surface-following methods estimate contact pose, shear, or contour features and use them for closed-loop servoing over rigid or geometrically structured surfaces~\cite{lepora2021poseservo,lloyd2024poseshear,velasco2026vbtmpc}. These methods provide important mechanisms for maintaining physical contact, but hair stroking introduces a different interaction regime. As hair fibers deform, slide, and redistribute beneath the end-effector, contact may remain present while its spatial location gradually shifts across the tactile surface. This drift is particularly difficult to observe using an external camera because the relevant interaction occurs directly underneath the end-effector. Consequently, successful trajectory progression, force regulation, or binary contact detection alone does not necessarily indicate that the robot is maintaining a spatially well-aligned stroking interaction.

\begin{figure}[t]
    \centering
    \includegraphics[width=\columnwidth]{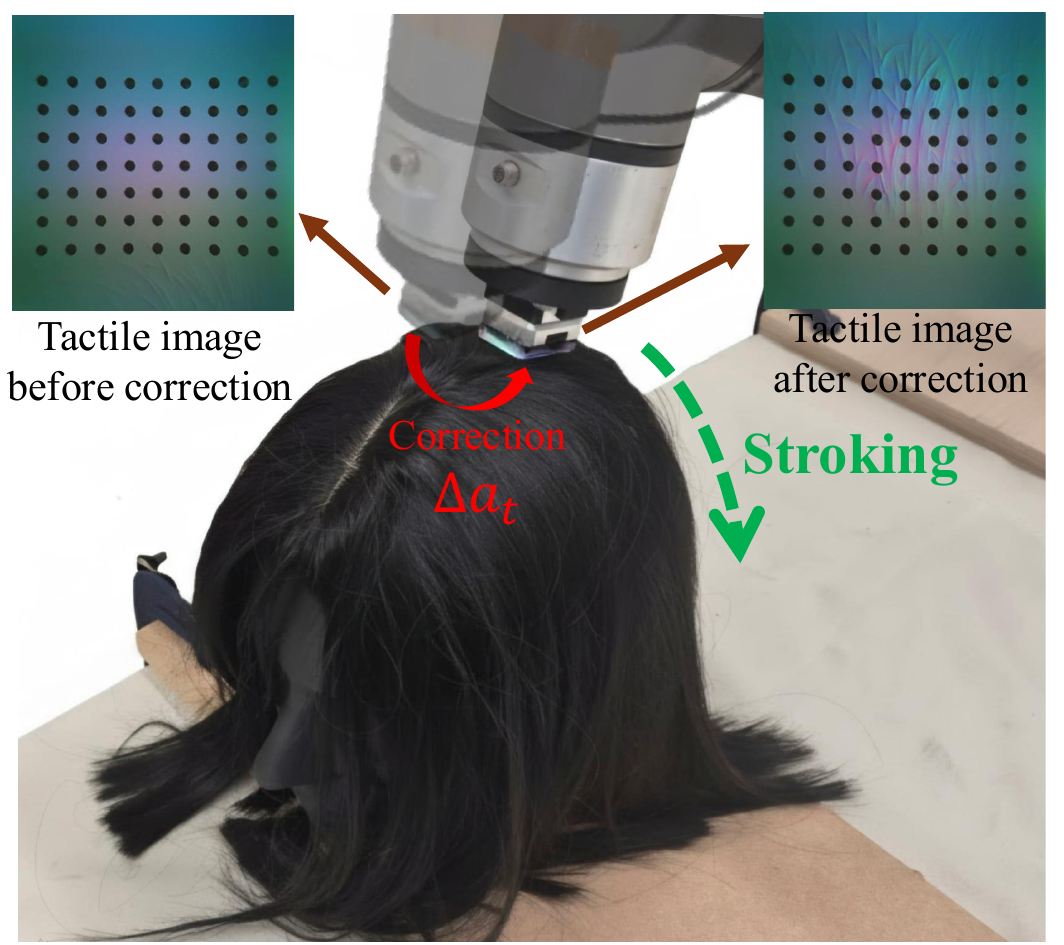}
    \caption{
    Concept of contact-distribution guided robotic hair stroking.
    A nominal visuotactile policy generates the overall stroking motion,
    while high-resolution tactile images provide explicit feedback for
    online residual correction $\Delta a_t$. The resulting joint command combines the
    nominal motion with the learned corrective residual.
    }
    \label{fig:01}
\end{figure}

To explicitly capture this local interaction, we represent sensor--hair contact as a spatial tactile contact distribution. High-resolution tactile observations are mapped to a compact representation describing the location and extent of hair-contact evidence over the sensing surface. To the best of our knowledge, prior robotic hair-care systems have not explicitly represented this spatial contact distribution as a feedback variable for learned online correction during continuous hair stroking.
The desired interaction is defined directly in tactile image space as a central contact region, avoiding the need to reconstruct the instantaneous 3D contact geometry of deformable hair. Building on this representation, we separate the control problem into \emph{nominal stroking} and \emph{contact recovery}. The nominal visuotactile policy is learned from complete expert demonstrations and generates the overall crown-to-back motion under near-nominal interaction. Because substantial contact deviations and recovery behaviors are sparse in such demonstrations, we additionally collect dedicated recentering data from laterally biased contacts and train a residual module to modify the nominal action only when the observed contact distribution leaves the desired region. This decomposition allows the nominal policy to focus on task progression while the correction module specializes in local contact recovery.

We evaluate the proposed framework in 525 real-robot trials across five head geometries and three hair conditions. Task success requires both sufficient stroke progression and contact maintenance. Compared with the same visuotactile policy without online correction, our method improves task success from 42.9\% to 62.3\% and contact maintenance from 59.4\% to 88.0\%, while maintaining a comparable stroke-completion rate. Continuous contact metrics further improve in 14 of the 15 tested headform--hair combinations. These results indicate that explicit contact recovery complements tactile conditioning within the nominal policy and that spatial tactile contact distributions provide a useful intermediate representation between high-dimensional tactile sensing and reactive control for deformable, visually occluded interactions.

In summary, the main contributions of this work are:
\begin{itemize}
    \item We formulate robotic hair stroking as a deformable surface-following problem in which physical contact can persist while spatially drifting, and introduce a compact tactile contact distribution to explicitly represent the location and extent of local hair--sensor interaction.
    \item We propose a contact-distribution guided online residual correction framework that separates nominal task progression from local contact recovery, using complete stroking demonstrations for nominal behavior and dedicated recentering demonstrations for corrective behavior.
    \item We demonstrate through 525 real-robot trials that the proposed correction substantially improves contact-preserving task success over vision-only and visuotactile imitation-learning baselines across diverse head geometries and hair conditions.
\end{itemize}

\section{Related Work}
\label{sec:related_work}

\subsection{Robotic Hair Interaction and Tactile Surface Following}

Robotic hair interaction has been studied for personal-care tasks including 
brushing, combing, detangling, and head care. Early head-care systems used 
compliant mechanisms and force regulation for scalp washing and 
massage~\cite{ando2013headcare}. Dennler \emph{et al.}~\cite{dennler2021haircombing} 
combined visual hair segmentation with trajectory planning for autonomous 
hair combing, while Hughes \emph{et al.}~\cite{hughes2021detangling} used 
brushing-force feedback to adapt stroke strategies during hair detangling. 
More recently, MOE-Hair~\cite{yoo2025moehair} introduced a soft compliant 
end-effector for contact-rich hair-care tasks such as head patting and finger 
combing, with interaction forces estimated from visual deformation and tendon 
measurements. These studies primarily address hair configuration change, 
compliant physical interaction, or force-aware manipulation. In contrast, 
we focus on maintaining spatially well-aligned contact throughout a continuous 
stroking motion as the hair--sensor interaction evolves.

Hair stroking is also closely related to tactile surface following. Prior 
work has estimated local contact pose from tactile observations for closed-loop 
servoing on unknown surfaces~\cite{lepora2021poseservo}, while pose-and-shear 
tactile servoing further accounts for tangential interaction and sensor 
deformation~\cite{lloyd2024poseshear}. More recently, VBT-MPC~\cite{velasco2026vbtmpc} 
directly regulates tactile contour features using model predictive control. 
These methods demonstrate the effectiveness of tactile feedback for contact 
regulation, but typically operate on rigid or geometrically structured 
interfaces. Hair introduces a deformable and continuously changing interaction 
layer in which contact can remain present while its spatial distribution 
gradually shifts across the sensor. Our work therefore uses the spatial 
distribution of tactile hair-contact evidence as an explicit feedback 
representation for detecting and correcting such contact deviations online.

\subsection{Visuotactile Policy Learning and Online Tactile Correction}

Visuotactile policy learning has shown that tactile sensing can complement 
vision in contact-rich manipulation by providing local physical information 
that is difficult to infer visually. Recent approaches integrate tactile 
observations into imitation or diffusion policies through multimodal feature 
fusion~\cite{huang2025vitac,helmut2026farm, zhao2026vitactracing}, while generalist manipulation 
models further incorporate tactile sensing into language-conditioned 
policies~\cite{huang2025tactilevla,zhang2026vtla}. These methods primarily use 
tactile information as part of the observation space for action generation.

A complementary line of work uses tactile feedback for reactive action 
adaptation. Reactive Diffusion Policy~\cite{xue2025reactive} combines a 
temporally coherent diffusion policy with a high-frequency tactile-reactive 
module, while VLA-Touch~\cite{bi2026vlatouch} uses a tactile-conditioned 
controller to refine actions produced by a pretrained VLA (Vision-Language-Action) policy. 
MimicTouch~\cite{yu2025mimictouch} similarly combines imitation learning with 
residual policy adaptation. Our work follows the general principle of 
separating nominal behavior generation from local reactive correction, but 
focuses on a task-specific feedback representation for deformable hair 
interaction. Rather than relying only on implicit tactile features, we 
explicitly represent hair--sensor interaction as a spatial contact distribution 
and use its deviation from a desired region to activate and condition learned 
residual corrections during stroking. Our distinction is not tactile conditioning or residual refinement itself, but the explicit spatial contact-distribution representation used to identify and recover contact drift during deformable hair interaction.

\section{Methodology}
\label{sec:methodology}

\begin{figure*}[t]
    \centering
    \includegraphics[width=\textwidth]{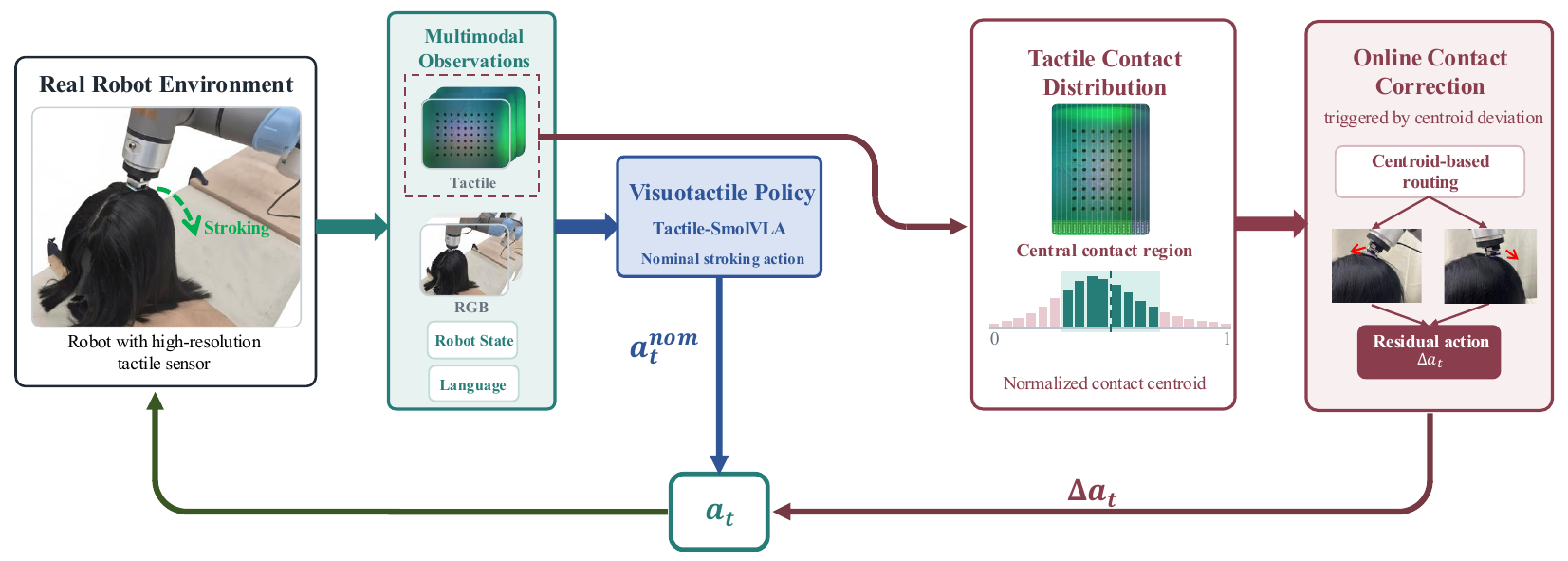}
    \caption{
    Overview of the proposed tactile contact-distribution guided
    hair-stroking framework. A nominal visuotactile policy takes RGB,
    tactile, robot-state, and language observations to generate
    absolute joint-position targets for the overall stroking motion.
    During execution, high-resolution tactile observations are mapped
    to a spatial hair-contact distribution, from which the normalized
    contact centroid is estimated. When the centroid deviates from the
    desired contact region, centroid-based routing activates the
    corresponding correction action, which predicts a short-horizon
    residual joint correction conditioned on the contact distribution
    and current robot state. The residual is added to the nominal joint
    target, and the resulting command is constrained and temporally
    smoothed before being sent to the robot. The nominal policy therefore
    drives stroke progression, while tactile feedback provides online
    local contact regulation.
    }
    \label{fig:method_pipeline}
\end{figure*}

\subsection{Task Formulation and System Overview}
\label{sec:task_overview}

As shown in Fig.~\ref{fig:01} and Fig.~\ref{fig:method_pipeline}, we consider robotic hair stroking as a sequential control problem in which the robot moves from the crown toward the back of a hair-covered head while maintaining stable local contact throughout the motion. At time step $t$, the robot observes an external RGB image $I_t^{v}$, a high-resolution tactile image $I_t^{\tau}$, and its joint state $q_t$. We denote the corresponding observation as $o_t=(I_t^{v},I_t^{\tau},q_t)$. The robot action $a_t$ is a six-dimensional absolute joint-position target. A fixed language instruction is additionally provided to the SmolVLA backbone, but it remains unchanged across all trials and is therefore omitted from the task formulation.

The task contains two complementary requirements: progressing along the prescribed stroking motion and maintaining the desired sensor--hair interaction. The nominal visuotactile policy predicts an action $a_t^{\mathrm{nom}}$ from $o_t$ to drive the overall crown-to-back motion. Meanwhile, the tactile observation is converted into a spatial contact distribution, whose centroid $\mu_t$ describes the lateral location of the current contact on the tactile surface. A predefined central region represents the desired contact location. When the estimated contact moves outside this region, the online correction module predicts a residual joint action $\Delta a_t$ to recenter the interaction; otherwise, $\Delta a_t=0$. The final command is therefore given by $a_t=\mathcal{S}(a_t^{\mathrm{nom}}+\Delta a_t)$, where $\mathcal{S}$ applies the shared joint limits, displacement constraints, and temporal smoothing. In this formulation, the nominal policy is responsible for stroke progression, while the correction module is responsible for recovering local contact deviation. At evaluation time, a successful trial is required to satisfy both stroke completion and contact maintenance, as defined in Section~\ref{sec:comparison_protocol}.

\subsection{Tactile Hair-Contact Distribution}
\label{sec:tactile_representation}

We represent local hair--sensor interaction by the spatial distribution of contact evidence over the tactile surface. The current and preceding high-resolution tactile frames are aligned, cropped, resized, background-subtracted, and concatenated before being encoded by a frozen Sparsh-DINO encoder~\cite{higuera2024sparsh} into a spatial feature grid $\mathbf{F}_t$. A lightweight decoder predicts hair-contact evidence over $K=20$ uniformly spaced bands along the lateral image axis:
\begin{equation}
    \mathbf{p}_t=\sigma\!\left(D_{\theta}(\mathbf{F}_t)\right), \qquad \mathbf{p}_t\in[0,1]^K,
    \label{eq:band_probabilities}
\end{equation}
where each $p_{t,i}$ denotes the probability of visible hair-contact traces in band $i$. Supervision is defined directly in tactile image space using manually annotated contact bands, rather than metric 3D contact points or forces. The decoder is trained with class-weighted binary cross-entropy while the tactile encoder remains frozen.

For a non-negligible total response, we normalize the band probabilities as
\begin{equation}
    w_{t,i}=\frac{p_{t,i}}{\sum_{j=0}^{K-1}p_{t,j}}, \qquad
    \mu_t=\sum_{i=0}^{K-1}\frac{i}{K-1}w_{t,i},
    \label{eq:contact_centroid}
\end{equation}
where $\mathbf{w}_t$ describes the relative spatial contact distribution and $\mu_t\in[0,1]$ is its normalized lateral centroid. Regional masses, spread, entropy, and the centroid are further extracted as compact distribution statistics for online correction.

\subsection{Nominal Visuotactile Stroking Policy}
\label{sec:visuotactile_policy}

We use a tactile-augmented SmolVLA~\cite{shukor2025smolvla} (tactile-SmolVLA) as the nominal stroking policy. Its role is to generate the overall crown-to-back motion under regular interaction, while explicit recovery from off-nominal contact is handled separately by the correction module in Section~\ref{sec:tactile_correction}. At each replanning step, the policy receives the external RGB observation, robot joint state, fixed task instruction, and tactile observations, and predicts a sequence of absolute joint-position targets.

The tactile frames are encoded by the same frozen Sparsh encoder used in Section~\ref{sec:tactile_representation}. Their spatial features are projected to the SmolVLA embedding space and aggregated into tactile tokens through a lightweight cross-attention adapter. We initialize from a stroke-finetuned SmolVLA checkpoint and retain its original conditional flow-matching objective~\cite{shukor2025smolvla,lipman2023flowmatching}, optimizing only the tactile adapter and action-related layers. We denote the current nominal action by $a_t^{\mathrm{nom}}$, which is executed directly when contact remains well aligned and otherwise modified by the online correction module.

\subsection{Contact-Distribution Guided Online Correction}
\label{sec:tactile_correction}

Complete stroking demonstrations predominantly contain near-nominal interactions and provide relatively few examples of substantial contact deviation and recovery. We therefore learn contact recovery from dedicated recentering demonstrations initialized with laterally biased contact. The desired interaction is defined by a central region on the tactile surface; when the predicted contact distribution leaves this region, the correction module generates a short-horizon joint-space residual $\Delta a_t$ to locally modify the nominal action $a_t^{\mathrm{nom}}$.

The correction input $\mathbf{x}_t\in\mathbb{R}^{32}$ contains the normalized contact distribution, six distribution statistics from Section~\ref{sec:tactile_representation}, and the six-dimensional robot joint state. The supervised target is
\begin{equation}
    \Delta a_t^{E}=a_{t+H}^{E,\mathrm{cmd}}-q_t, \qquad H=5,
    \label{eq:expert_correction_target}
\end{equation}
where $a_{t+H}^{E,\mathrm{cmd}}$ is the expert joint-position command approximately $0.5$\,s ahead. Thus, the annotated tactile data supervise the contact-distribution predictor, whose output determines the spatial contact state and whether correction is required, while the recentering demonstrations supervise how the robot should correct it.

We use two correction branches for the two possible centroid-recovery directions. Demonstration transitions are assigned to the centroid-increasing or centroid-decreasing branch according to the observed centroid change. A shared MLP with two six-dimensional output heads predicts the corresponding residual correction and is trained using a Smooth L1 loss. During deployment, the contact distribution is temporally filtered before correction. Contact within the desired central region leaves the nominal action unchanged, while deviations activate the corresponding recovery branch. The predicted correction is further scaled with contact deviation and clipped to joint-wise limits. Let $\widehat{\Delta a}_t$ denote the predicted correction, $b_t$ the activation gate, and $\gamma_t$ the deviation-dependent gain. The applied correction and final action are
\begin{equation}
\begin{aligned}
    \Delta a_t &= b_t\,\operatorname{clip}\!\left(\gamma_t\widehat{\Delta a}_t,-\mathbf{c},\mathbf{c}\right),\\
    a_t &= \mathcal{S}\!\left(a_t^{\mathrm{nom}}+\Delta a_t\right),
\end{aligned}
\label{eq:online_tactile_correction}
\end{equation}
where $\mathbf{c}$ denotes the joint-wise correction limits and $\mathcal{S}$ applies the shared execution constraints and temporal smoothing. The correction is zero when contact remains within the desired region. Importantly, the contact-quality metric introduced below is used only for evaluation and is not directly optimized by the controller.

\subsection{Evaluation Metrics}
\label{sec:evaluation_metrics}

Robotic hair stroking requires both task progression and contact preservation. Following related hair-care and tactile surface-following studies that evaluate task effectiveness together with physical interaction or contact tracking~\cite{dennler2021haircombing,yoo2025moehair,velasco2026vbtmpc}, we evaluate these two aspects separately and combine them for trial-level task success.

For contact preservation, let $\mathcal{C}$ denote the desired central region, $c_t$ its active-band coverage, and $\ell_t$ the longest contiguous active span intersecting $\mathcal{C}$. We define the frame-level contact-distribution quality as
\begin{equation}
    Q_t=c_t\left[\alpha+(1-\alpha)\min\left(\frac{\ell_t}{L_{\mathcal{C}}},1\right)\right],
    \label{eq:contact_quality}
\end{equation}
where central coverage penalizes lateral contact loss and the contiguous span characterizes contact continuity. For a quality threshold $\tau$, the time-weighted qualified-contact fraction is
\begin{equation}
    F_Q(\tau)=\frac{\sum_{t\in\mathcal{V}}\omega_t\,\mathbb{I}[Q_t\geq\tau]}{\sum_{t\in\mathcal{V}}\omega_t},
    \label{eq:qualified_contact_fraction}
\end{equation}
where $\mathcal{V}$ denotes valid tactile observations and $\omega_t$ is the temporal weight. We additionally report the time-weighted mean contact quality and \emph{Tail-Q}. Lateral alignment is measured by the instantaneous centroid error $e_t=|\mu_t-0.5|$ and, when needed, its time-weighted mean $\bar e$ over valid contact observations.

Stroke progression is evaluated in joint space. Let $\mathbf{q}^{\mathrm{arm}}_0$, $\mathbf{q}^{\mathrm{arm}}_t$, and $\mathbf{q}^{\mathrm{arm}}_g$ denote the initial, current, and terminal five-dimensional arm configurations, excluding the gripper. The normalized progress is
\begin{equation}
    P_t=\operatorname{clip}\left(1-\frac{\|\mathbf{q}^{\mathrm{arm}}_t-\mathbf{q}^{\mathrm{arm}}_g\|_2}{\|\mathbf{q}^{\mathrm{arm}}_0-\mathbf{q}^{\mathrm{arm}}_g\|_2},0,1\right),
    \label{eq:stroke_progress}
\end{equation}
and $P_{\mathrm{tail}}$ is the median progress over the final evaluation interval. A trial is considered successful only when both the stroke-completion and contact-maintenance criteria are satisfied; their fixed thresholds are specified in Section~\ref{sec:comparison_protocol}. The tactile metrics are image-based proxies rather than measurements of physical force or comfort. Although the controller and evaluator share the same frozen tactile representation, $Q_t$ is computed offline and is not directly optimized by the correction controller.

\section{Experiments}
\label{sec:experiments}

\subsection{Experimental Setup and Training Protocol}
\label{sec:experimental_setup}

\begin{figure}[t]
    \centering
    \includegraphics[width=\columnwidth]{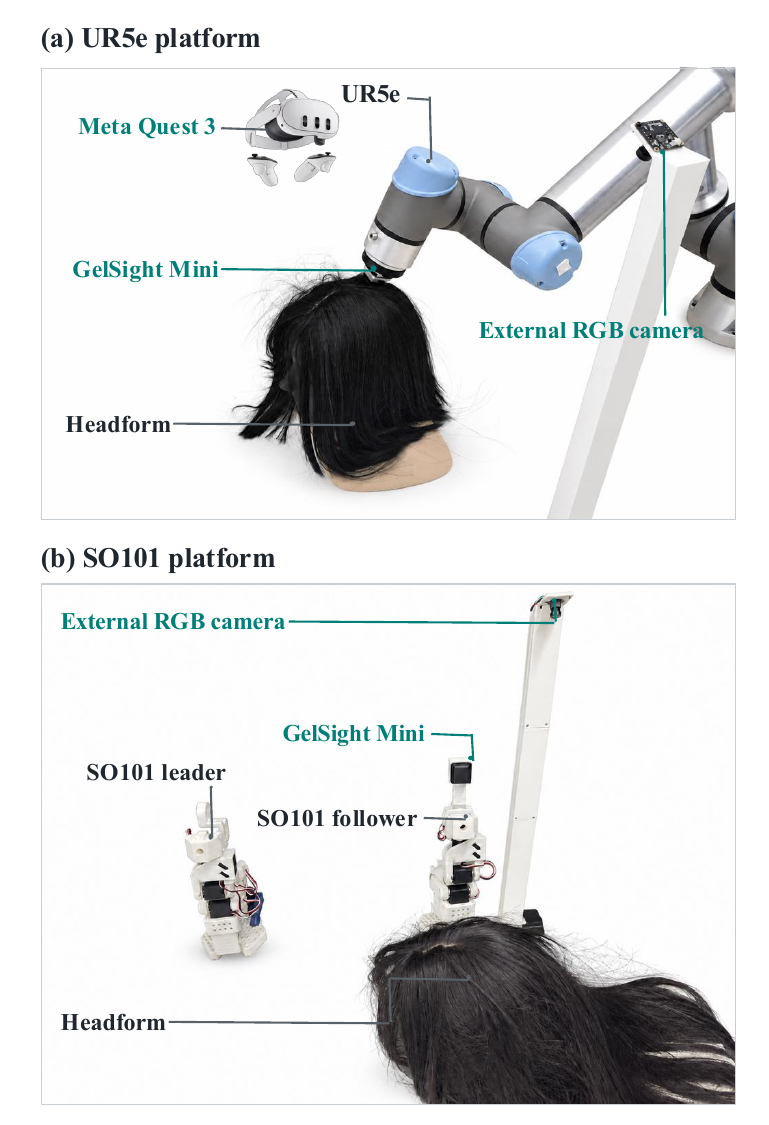}
    \caption{Experimental platforms. (a) The UR5e platform comprises
    a UR5e robot, a Meta Quest 3 teleoperation interface~\cite{alian2026haptile}, an external
    RGB camera, and an end-effector-mounted GelSight Mini.
    (b) The SO101 platform comprises an SO101 leader--follower pair~\cite{knight2024so101},
    the same external RGB camera used in the UR5e platform, and a
    GelSight Mini mounted on the follower end effector. Both platforms
    interact with a secured hair-covered headform. The quantitative
    experiments reported in this work were conducted using the SO101
    platform.}
    \label{fig:exp_platform}
\end{figure}

Our experimental system consists of perception, robotic, and teleoperation components, as shown in Fig.~\ref{fig:exp_platform}. A GelSight Mini tactile sensor is mounted at the robot end effector, while a fixed third-person RGB camera provides global visual observations at $640\times480$ resolution. We evaluate both the standard GelSight Mini membrane and a variant equipped with a custom silicone membrane. The SO101 serves as the primary platform for quantitative evaluation, while the UR5e is used for additional cross-platform validation. Because the two robots have different kinematics and joint spaces, their nominal policies and correction modules are trained separately. Teleoperated demonstrations are collected using an SO101 leader or Meta Quest 3, with real-time tactile observations displayed to the operator.

\begin{figure}[t]
    \centering
    \includegraphics[width=\linewidth]{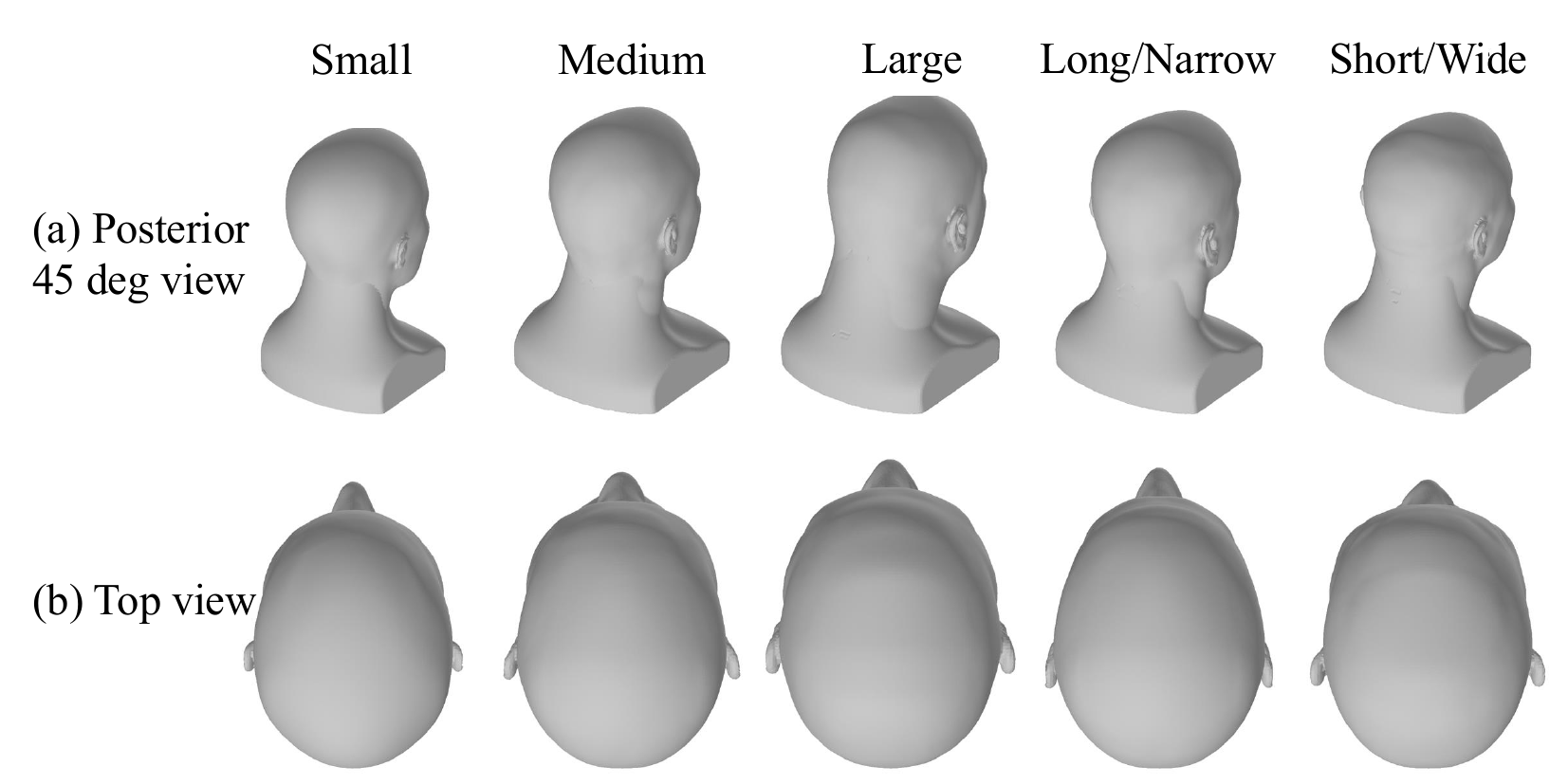}
    \caption{Five CDC/NIOSH headforms~\cite{niosh2020headforms} used for evaluation. All models are printed at their original scale.}
    \label{fig:headforms}
\end{figure}

\begin{figure}[t]
    \centering
    \includegraphics[width=\columnwidth]{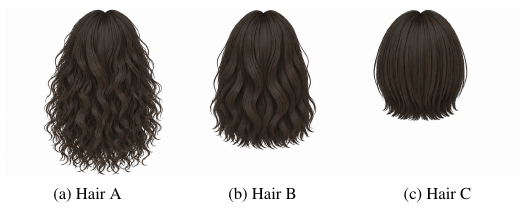}
   \caption{Schematic illustrations of the three evaluated hair
conditions. Hair A, Hair B, and Hair C differ jointly in length,
curl, and smoothness and are treated as distinct physical
conditions rather than controlled variations of a single property.
The illustrations provide qualitative visualizations and are not
photographs or physical measurements of the wigs.}
    \label{fig:hair_conditions}
\end{figure}

We evaluate five standardized CDC/NIOSH headforms~\cite{niosh2020headforms}: Small, Medium, Large, Long/Narrow, and Short/Wide, together with three hair conditions varying jointly in length, curl, and smoothness, as shown in Figs.~\ref{fig:headforms} and~\ref{fig:hair_conditions}. Training data are collected using the Small headform with Hair A. The nominal stroking dataset contains 97 demonstrations and 7,002 frames; the contact-distribution predictor is trained from 1,045 manually annotated tactile images; and the correction module uses 75 dedicated recentering demonstrations initialized with laterally biased contact. The nominal policy, contact predictor, and correction module are trained separately using their corresponding training/validation splits. Unless otherwise specified, all quantitative results are obtained on the SO101 platform.

\subsection{Compared Methods and Evaluation Protocol}
\label{sec:comparison_protocol}

We compare three configurations. \textit{SmolVLA} uses RGB observations, robot states, and the fixed task instruction without tactile input. \textit{Tactile-SmolVLA} additionally conditions the nominal policy on GelSight features but does not perform explicit online correction. \textit{Ours} uses the same Tactile-SmolVLA checkpoint together with the contact-distribution-guided correction module in Section~\ref{sec:tactile_correction}. The comparison between Tactile-SmolVLA and Ours therefore isolates the effect of explicit online correction.

The main evaluation comprises 525 SO101 trials over all 15 combinations of five headforms and three hair conditions. Each method is evaluated in 175 trials: 15 repetitions per headform for Hair A and 10 repetitions for Hair B and Hair C. All methods share the same initial configuration, terminal criterion, task instruction, replanning rate, joint limits, temporal smoothing, and execution constraints. Before each trial, tactile contact is initialized near the center of the sensor, after which execution proceeds without physical guidance.

All trial-level metrics are computed within a common 45\,s analysis window. Stroke completion requires $P_{\mathrm{tail}}\geq0.75$, while contact maintenance requires $F_Q(0.50)\geq0.40$; task success requires both criteria. The continuous qualified-contact fraction is separately reported as $F_Q(0.60)$. All thresholds are fixed across methods. Confidence intervals are obtained by bootstrap resampling within each condition, while the episode-level analysis in Section~\ref{sec:correction_analysis} resamples correction episodes.

\subsection{Results Across Headforms and Hair Conditions}
\label{sec:cross_condition_results}

\begin{figure}[t]
    \centering
    \includegraphics[width=\columnwidth]{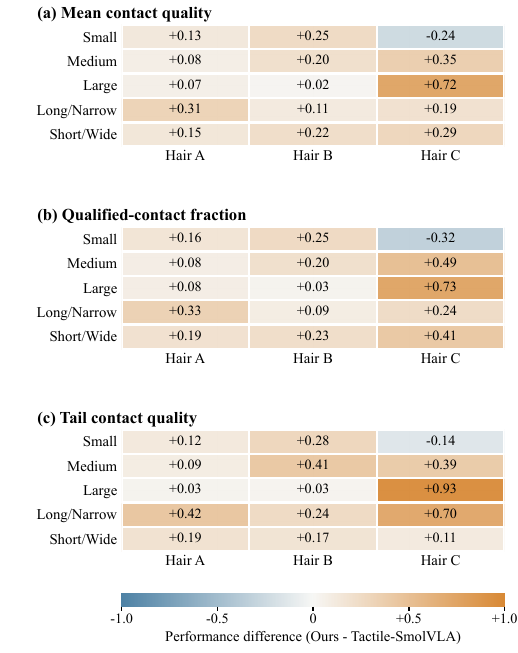}
    \caption{Condition-wise effect of online tactile correction across five headforms and three hair conditions. Each cell reports the difference between Ours and Tactile-SmolVLA in (a) mean contact quality, (b) qualified-contact fraction $F_Q(0.60)$, and (c) Tail-Q. Positive values indicate improvement from online correction, while negative values indicate reduced performance.}
    \label{fig:main_results}
\end{figure}

\begin{table}[t]
    \centering
    \caption{Trial-level success rates across 525 trials, comprising 175 trials per method. Stroke completion requires $P_{\mathrm{tail}}\geq0.75$, contact maintenance requires $F_Q(0.50)\geq0.40$, and task success requires both criteria.}
    \label{tab:main_results}
    \footnotesize
    \setlength{\tabcolsep}{1.5pt}
    \renewcommand{\arraystretch}{1.08}
    \begin{tabular}{@{}lccc@{}}
        \hline
        \multicolumn{1}{l}{\textbf{Method}} &
        \shortstack[c]{\textbf{Stroke completion}\\\textbf{rate}} &
        \shortstack[c]{\textbf{Contact maintenance}\\\textbf{rate}} &
        \shortstack[c]{\textbf{Task success}\\\textbf{rate}} \\
        \hline
        SmolVLA & 73.1\% & 51.4\% & 36.6\% \\
        Tactile-SmolVLA & 74.3\% & 59.4\% & 42.9\% \\
        Ours & 72.6\% & 88.0\% & \textbf{62.3\%} \\
        \hline
        
    \end{tabular}
\end{table}

\begin{figure}[t]
    \centering
    \includegraphics[width=\columnwidth]{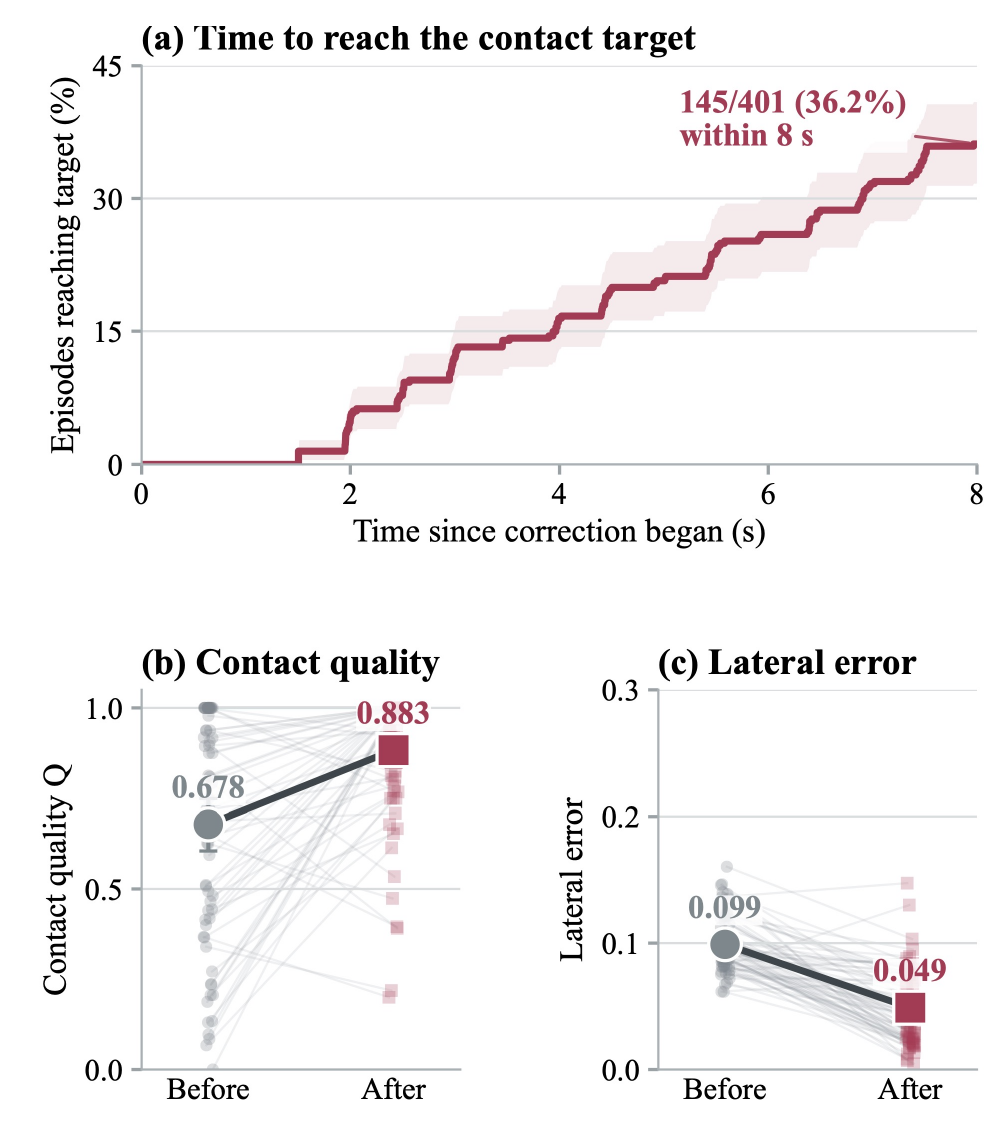}
    \caption{Episode-level analysis of online tactile correction. (a) Cumulative fraction of correction episodes reaching the contact target within 8\,s of the first applied residual action, with pointwise 95\% bootstrap confidence intervals. Among 401 episodes with complete observation windows, 145 reach the target, with a median confirmation time of 4.432\,s. (b--c) Contact quality and time-weighted mean lateral error before correction and after confirmed target attainment for 70 confirmed target-reaching episodes with complete 2\,s observation windows.}
    \label{fig:correction_analysis}
\end{figure}

\begin{figure*}[t]
    \centering
    \includegraphics[width=\textwidth]{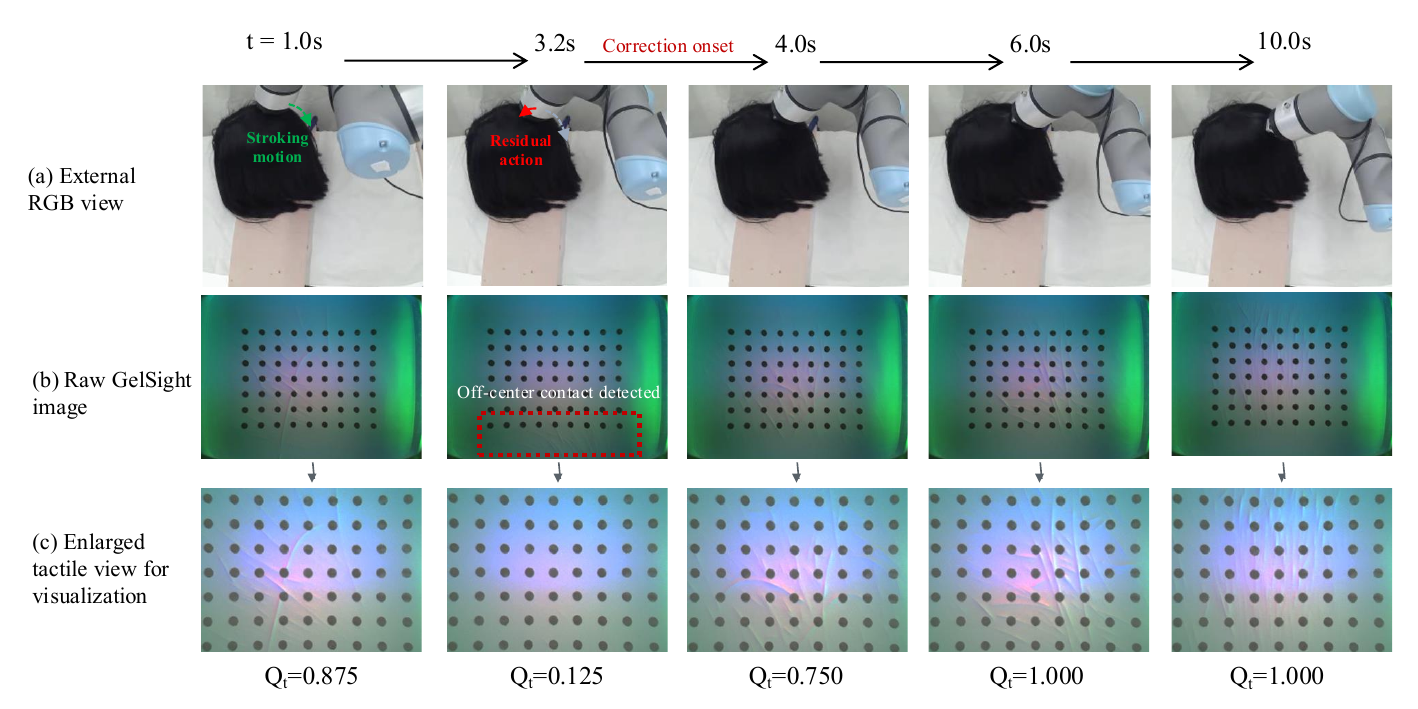}
    \caption{Representative sequence from a UR5e stroking rollout
using the proposed contact-distribution-guided online correction.
The columns show synchronized external RGB and GelSight
observations at the indicated times after stroke onset. At
$t=3.2$\,s, the system detects off-center contact and initiates
a residual correction (red arrow) while the nominal stroking
motion continues (blue dashed arrows). Contact quality subsequently
increases from $Q_t=0.125$ to $0.750$ and then to $1.000$.
Row (c) shows enlarged versions of the corresponding tactile
images for visualization only; these enlarged views are not
additional model inputs.}

    \label{fig:qualitative_stroking}
\end{figure*}

Table~\ref{tab:main_results} reports the trial-level results. Ours achieves a task-success rate of 62.3\%, compared with 42.9\% for Tactile-SmolVLA and 36.6\% for SmolVLA. Relative to Tactile-SmolVLA, this corresponds to an improvement of 19.4 percentage points, with a 95\% bootstrap confidence interval of $[11.4,\,27.4]$. The main gain arises from contact preservation: contact-maintenance rate increases from 59.4\% to 88.0\%, while stroke-completion rate remains comparable (74.3\% versus 72.6\%). Continuous metrics show the same trend, with mean contact quality increasing from $0.459$ to $0.649$, $F_Q(0.60)$ from $0.422$ to $0.635$, and Tail-Q from $0.480$ to $0.746$.

Fig.~\ref{fig:main_results} shows the condition-wise changes relative to Tactile-SmolVLA. Ours improves all three continuous contact metrics in 14 of the 15 headform--hair combinations. The only consistent negative case is Small--Hair-C, where mean contact quality, qualified-contact fraction, and Tail-Q decrease by $0.237$, $0.319$, and $0.139$, respectively. Since correction remains beneficial for the same headform with the other hair conditions and for Hair C with the other headforms, this failure appears to depend on the combined local geometry and hair condition rather than either factor alone.

We further instantiate the same contact-distribution-guided framework on a UR5e manipulator and with both standard and custom tactile membranes. Because the UR5e has different kinematics and joint space, its nominal policy and correction module are retrained rather than directly transferred from the SO101. These qualitative experiments therefore evaluate framework-level applicability across robot embodiments and tactile interfaces rather than zero-shot policy generalization.

\subsection{Analysis of Online Tactile Correction}
\label{sec:correction_analysis}

We identify 483 non-overlapping correction episodes during evaluation. Among the 401 episodes with a complete 8\,s observation window, 145 (36.2\%) reach the predefined contact target, which requires valid tactile contact, $Q_t\geq0.60$, and instantaneous lateral error $e_t\leq0.08$ for three consecutive observations. The median confirmation time among target-reaching episodes is 4.432\,s.

For 70 confirmed episodes with complete 2\,s windows before correction onset and after confirmation, mean contact quality increases from $0.678$ to $0.883$, while the time-weighted mean lateral error $\bar e$ decreases from $0.099$ to $0.049$. Fig.~\ref{fig:correction_analysis} summarizes these dynamics. Because the target-attainment criterion is stricter than detecting incremental improvement, episodes that do not reach this target should not be interpreted as receiving no benefit; the trial-level results in Table~\ref{tab:main_results} remain the primary measure of overall task performance.

\subsection{Qualitative Stroking Sequence}
\label{sec:qualitative_stroking}

Fig.~\ref{fig:qualitative_stroking} shows a representative
UR5e stroking sequence with synchronized external RGB and GelSight
observations. The sequence qualitatively demonstrates continued
stroke progression together with online contact recentering under
a robot embodiment and sensor mounting different from those of
the primary SO101 platform. It complements the SO101 quantitative
evaluation by relating global robot motion to the evolving local
hair--sensor interaction.

\subsection{Discussion and Limitations}
\label{sec:discussion}

The results indicate that explicit contact-distribution-guided recovery complements tactile conditioning in the nominal policy, substantially improving contact maintenance without materially changing stroke completion. However, the Small--Hair-C failure case suggests that a fixed correction strategy may not suit all combinations of local geometry and hair properties, motivating adaptive correction magnitude or activation. The contact metrics also share the same frozen tactile representation used by the controller and therefore are not independent physical measurements, although the reported contact-quality score is not directly optimized during control. Moreover, the tested wigs vary jointly in multiple properties, and the main experiments use stationary headforms and a prescribed crown-to-back stroke. Evaluation on moving human heads, additional stroking directions, and direct measurements of contact force and human comfort remain important future directions.

\section{Conclusion}
\label{sec:conclusion}

We presented a tactile contact-distribution guided online correction framework for robotic hair stroking. The proposed method represents local hair--sensor interaction as a spatial tactile contact distribution and uses this representation to detect contact deviation and guide short-horizon residual corrections on top of a nominal visuotactile stroking policy. Across 525 SO101 trials covering five headforms and three hair conditions, our method improves task success from 42.9\% to 62.3\% and contact maintenance from 59.4\% to 88.0\% compared with the same visuotactile policy without online correction, while maintaining a comparable stroke-completion rate. Continuous contact metrics further improve in 14 of the 15 tested headform--hair combinations. Additional qualitative validation on a UR5e platform and with an alternative tactile membrane further illustrates the applicability of the framework beyond the primary SO101 setup. More broadly, the results suggest that deformable contact can be represented not only by force or binary contact state, but also by its spatial distribution as an intermediate variable for reactive manipulation. Future work will investigate adaptive correction strategies, moving heads and additional stroking directions, and independent measurements of contact force and human comfort.

\bibliographystyle{IEEEtran}
\bibliography{reference}

\end{document}